%% file: main.tex
\documentclass{article}

\usepackage{iclr2027_conference,times}

\usepackage{amsmath,amssymb}
\usepackage{graphicx}
\graphicspath{{figure/}}
\usepackage{booktabs}
\usepackage{tabularx,array}
\usepackage{float}
\usepackage{placeins}
\usepackage{xcolor}
\usepackage[protrusion=true,expansion=false,patch=none]{microtype}
\usepackage[unicode=true,colorlinks=false,
            pdfauthor={Pengan Chen, Liang Hong, Xinyuan Liu, Han Huang, Ziqian Lin,
                       Jiayang Chen, Yimin Fan, Jiyue Jiang, Yu Li},
            pdftitle={Scaling RNA foundation models through iterative refinement}]{hyperref}

\title{Scaling RNA foundation models through \\ iterative refinement}

\author{%
\bfseries Pengan Chen$^{*}$ \quad Liang Hong$^{*}$ \quad Xinyuan Liu$^{*}$ \quad Han Huang \quad Ziqian Lin \\
\bfseries Jiayang Chen \quad Yimin Fan \quad Jiyue Jiang \quad Yu Li$^{\dagger}$ \\[6pt]
{\normalfont\small Department of Computer Science and Engineering, The Chinese University of Hong Kong} \\
{\normalfont\small\texttt{\{chenpengan,liang.hong,xinyuan\}@link.cuhk.edu.hk}, \texttt{liyu@cse.cuhk.edu.hk}}%
}

\iclrfinalcopy

\begin{document}

\maketitle
{\renewcommand{\thefootnote}{\ensuremath{\fnsymbol{footnote}}}%
\footnotetext[1]{Equal contribution.}%
\footnotetext[2]{Corresponding author.}}
\lhead{Technical Report}
\pagestyle{fancy}
\thispagestyle{fancy}

\begin{abstract}
\input{sections/abstract}
\end{abstract}

\section{Introduction}
\label{sec:intro}
\input{sections/introduction}

\section{Method}
\label{sec:method}
\input{sections/method}

\FloatBarrier
\section{Experiments}
\label{sec:experiments}
\input{sections/experiments}

\section{Discussion and conclusion}
\label{sec:discussion}
\input{sections/discussion}

\input{sections/statements}

\bibliographystyle{iclr2027_conference}
\bibliography{reference}

\clearpage
\appendix
\setcounter{table}{0}\renewcommand{\thetable}{A\arabic{table}}\renewcommand{\theHtable}{A\arabic{table}}
\setcounter{figure}{0}\renewcommand{\thefigure}{A\arabic{figure}}\renewcommand{\theHfigure}{A\arabic{figure}}
\input{sections/appendix}

\FloatBarrier
\section{Related work}
\label{sec:related}
\label{app:related}
\input{sections/related_work}

\end{document}

%% file: sections/abstract.tex
Biological function in RNA is encoded across scales, from local sequence motifs to long-range structural dependencies.
Conventional RNA foundation models couple parameter capacity and inference depth within a fixed architecture, prescribing a fixed computation graph for each inference.
Here we present Eco-RNA, an 80M-parameter RNA foundation model that makes inference depth adjustable through learned iterative refinement.
Its shared, input-conditioned Transformer core refines sequence representations through repeated application.
The iteration count is selected after training and controls effective depth at fixed model parameters.
We train across recurrent depths with masked reconstruction, state regularization, and a one-sided loss that anchors deeper reconstruction to a shallow reference.
On the 70-assay RNAGym fitness benchmark, increasing refinement from one to four iterations raises aggregate absolute Spearman correlation from 0.242 to 0.286, with the gain retained through eight iterations.
At four iterations, Eco-RNA surpasses the current SOTA Evo~2 40B on all three aggregate fitness metrics despite 500$\times$ fewer parameters.
Its iteration configurations advance the compute--performance Pareto frontier among the evaluated models.
It also outperforms the evaluated baselines in non-coding RNA family classification, RNA secondary-structure prediction, and chemical probing profile prediction.
These results establish internal inference computation as an effective scaling dimension for RNA foundation models.

%% file: sections/introduction.tex
RNA function depends on information distributed across a sequence, from local recognition motifs to interactions between distant nucleotides.
Foundation models learn transferable representations of this information from unannotated sequences and support the prediction of RNA structure and function~\citep{chen2022interpretable,penic2025rinalmo,zou2024aido}.
Progress has accompanied increases in model capacity, pretraining data, and context length~\citep{penic2025rinalmo,zou2024aido,brixi2026genome}.
Conventional backbones, however, apply a fixed stack of transformations at inference.
Each trained checkpoint therefore offers limited flexibility to trade additional computation for better predictions.

Internal test-time scaling makes execution depth an additional resource for prediction.
In language modeling, inference computation can improve predictions through extended generation or candidate selection~\citep{brown2024large,snell2024scaling}.
Recurrent-depth architectures realize internal scaling by repeatedly applying a shared transformation to a latent representation~\citep{Dehghani2018,geiping2025scaling,alabdulmohsin2025recursive}.
For a nucleotide encoder, this provides a direct way to allocate additional computation while preserving the sequence input and prediction interface.
The central learning challenge is to make the shared update predictive across computational budgets: it must retain access to sequence information, control the scale of the recurrent state, and produce useful representations at different depths.

Here, we introduce \textbf{Eco-RNA}, an 80M-parameter RNA foundation model trained for iterative refinement (Figure~\ref{fig:architecture}).
Building on recurrent-depth architectures~\citep{geiping2025scaling,prairie2026parcae}, Eco-RNA reuses an input-conditioned Transformer core with an iteration count chosen at inference.
Persistent input injection supplies sequence information at each iteration, while bounded state retention and normalization regulate the recurrent update.
Pretraining combines masked reconstruction at sampled terminal depths with a scale-normalized terminal-state penalty inspired by \citet{viakhirev2026think}.
Our one-sided reconstruction anchor adds supervision when an endpoint has higher reconstruction loss than a shallower reference on the same trajectory.
This formulation trains iterative computation within a self-supervised sequence backbone; RNAformer and RhoFold+ apply recurrence within task-specific structure predictors~\citep{franke2023scalable,shen2024accurate}.

\begin{figure}[t]
\centering
\includegraphics[width=0.8\textwidth]{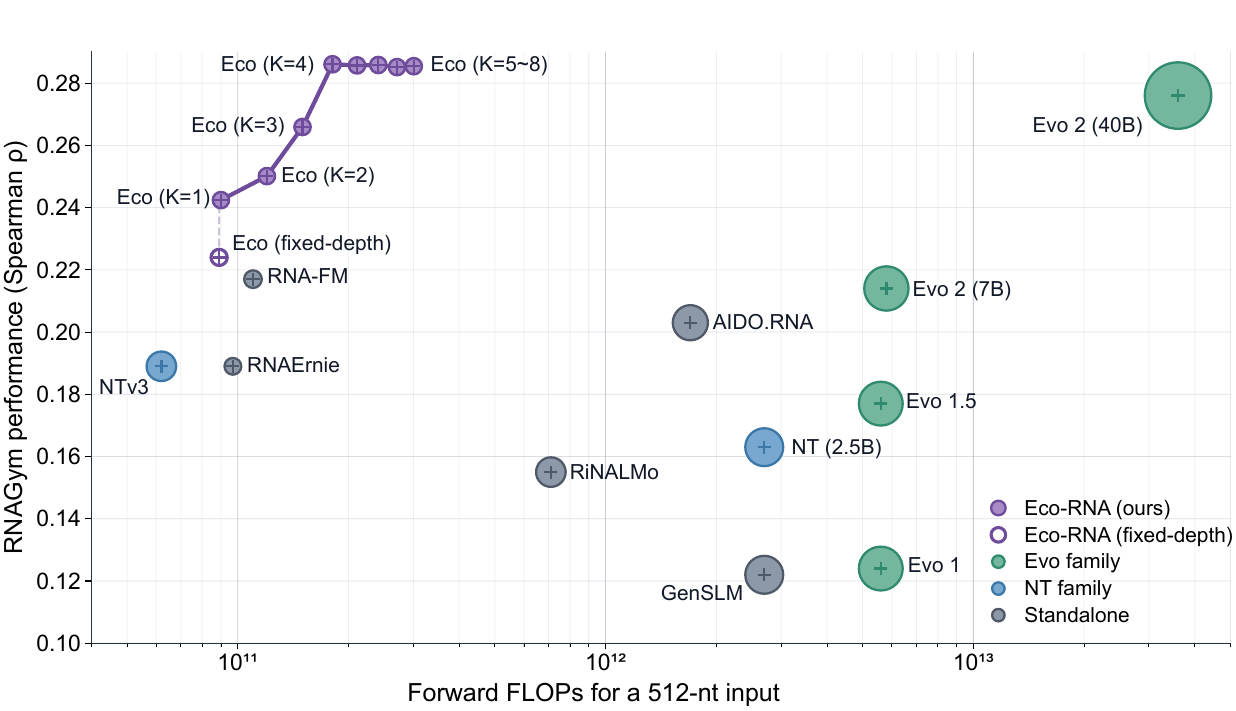}
\caption{\textbf{Inference compute versus RNA fitness prediction.}
Estimated FLOPs for one forward pass at 512 nucleotides (log scale) and type-balanced
aggregate absolute Spearman correlation on RNAGym. Colors identify model
families; marker size increases nonlinearly with parameter count. The line connects
$K{=}1,\ldots,8$ of one Eco-RNA checkpoint, and the hollow marker denotes the
fixed-depth control. FLOP accounting is given in Appendix~\ref{app:compute}.}
\label{fig:pareto}
\end{figure}

Eco-RNA advances the Pareto frontier of estimated forward FLOPs and RNA fitness prediction performance among the evaluated models (Figure~\ref{fig:pareto}).
Under the original 70-assay RNAGym protocol~\citep{arora2025rnagym}, its one- to four-iteration configurations provide four Pareto-optimal operating points from one checkpoint.
Increasing depth from one to four iterations raises aggregate absolute Spearman correlation from 0.242 to 0.286, an 18\% relative improvement at approximately twice the forward FLOPs.
The gain is retained through eight iterations.
At four iterations, Eco-RNA exceeds Evo~2 40B on all three aggregate fitness metrics with approximately 500$\times$ fewer parameters~\citep{brixi2026genome}.

The depth benefit also appears in held-out reconstruction across pretraining checkpoints and persists after supervised task adaptation.
Task-specific models initialized from Eco-RNA achieve the highest family-classification accuracy, secondary-structure F1, and probe-mean chemical-reactivity AUC among the compared models.
Within each task, the adapted checkpoint and prediction head remain fixed throughout the depth sweep.
The resulting curves exhibit different saturation depths, showing that a shared computational interface supports distinct prediction tasks while allowing their inference budgets to be chosen separately.

Our contributions are:
\begin{itemize}
\item \textbf{Internal test-time scaling for RNA foundation models.} Eco-RNA learns an input-conditioned shared update that supports adjustable inference depth within an 80M-parameter sequence backbone.
\item \textbf{Cross-depth training with one-sided reconstruction anchoring.} The anchor selectively reinforces deeper predictions relative to a reference loss, complementing supervision at sampled depths and terminal-state regularization.
\item \textbf{An improved compute--performance frontier.} Eco-RNA advances the RNAGym frontier through adjustable recurrence. Fixed-weight evaluations also establish depth-dependent gains in pretraining reconstruction and supervised RNA prediction.
\end{itemize}
Appendix~\ref{app:related} reviews related work on recurrent computation and biological foundation models.

%% file: sections/method.tex
\subsection{Recurrent depth as an inference budget}
\label{sec:method_budget}

Eco-RNA exposes inference depth as an explicit control over internal
computation. Its recurrent depth $K$ is the number of times a shared
Transformer core is applied; increasing $K$ allocates more computation to the
same input without changing the learned parameters or prediction interface. A
prelude encodes the sequence once, the shared core iteratively updates a latent
state, and a coda maps the state selected at depth $K$ to a prediction.

Let $x=(x_1,\ldots,x_L)$ be a tokenized nucleotide sequence, $\tilde x$ its
corrupted input, and $\mathcal M$ the positions selected for reconstruction.
Let $\mathcal V$ denote the vocabulary, $d$ the hidden width, and $\Theta$ all
model parameters. Eco-RNA comprises a prelude $P$, an input-conditioned
shared core $R_\theta$, and a coda $C$, with
$L_{\mathrm p}$, $L_{\mathrm r}$, and $L_{\mathrm c}$ Transformer blocks,
respectively; $\theta\subset\Theta$ denotes the parameters of the shared core.
The prelude and coda each execute once, whereas the core executes $K$ times, so
\begin{equation}
D(K)=L_{\mathrm p}+K L_{\mathrm r}+L_{\mathrm c}
\label{eq:effective_depth}
\end{equation}
Transformer blocks are applied. Thus, recurrent depth changes the executed
computation while leaving parameter count fixed~\citep{Dehghani2018,geiping2025scaling}.
Architecture values and the
corresponding compute model appear in Appendices~\ref{app:architecture}
and~\ref{app:compute}.

\begin{figure}[tbp]
\centering
\includegraphics[width=\textwidth]{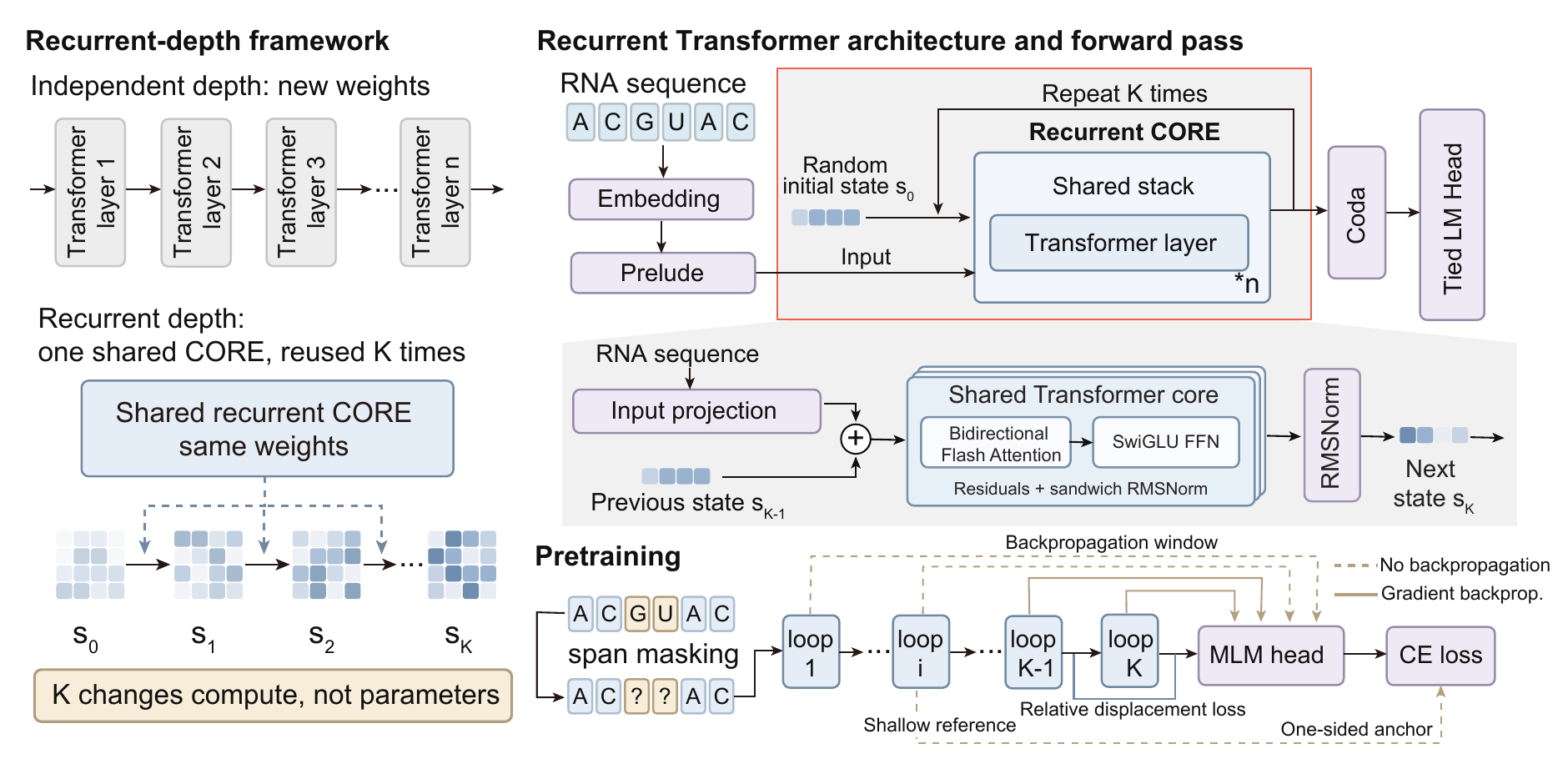}
\caption{\textbf{Eco-RNA.} Left: a conventional stack assigns distinct
parameters to successive layers, whereas Eco-RNA applies one shared core for
$K$ iterations, decoupling executed depth from parameter count. Right, top: a
prelude encodes the input once; the recurrent core updates a randomly
initialized state for $K$ iterations; and the coda with a tied language-model
head reads out the terminal state. Right, middle: each update combines retained
state with the projected prelude representation, applies the shared
Transformer core, and normalizes the next state. Bottom: pretraining applies
masked reconstruction at each sequence's sampled terminal depth, relative
displacement regularization to the final two states, and a one-sided
reconstruction anchor that compares eligible deep endpoints with a detached
reference.}
\label{fig:architecture}
\end{figure}

\subsection{An input-conditioned shared update}
\label{sec:method_recurrence}

Each recurrent update must combine the original sequence information with the
result of preceding computation. Eco-RNA supplies a fixed prelude
representation at every iteration and maintains a separate state that evolves
along the trajectory. With embedding matrix $E\in\mathbb R^{|\mathcal V|\times d}$ and
prelude normalization $N_{\mathrm p}$, these two inputs are initialized as
\begin{equation}
 e=N_{\mathrm p}\!\left(P(E(\tilde x))\right),
 \qquad
 s^{(0)}\sim\mathcal N(0,\sigma_{\mathrm{init}}^2 I),
\label{eq:state_initialization}
\end{equation}
where $e,s^{(k)}\in\mathbb R^{L\times d}$ and the hyperparameter
$\sigma_{\mathrm{init}}$ controls the initial-state scale. Padding entries of
the sampled state are set to zero.

Eco-RNA bounds the retained fraction of each state channel while learning the
scale of persistent input injection. Learnable channel vectors
$a,b\in\mathbb R^d$ parameterize a positive input scale $\delta$ and a
retention coefficient $\alpha$:
\begin{equation}
\begin{gathered}
\delta=\operatorname{softplus}(b),\\
\alpha=\alpha_{\min}+(\alpha_{\max}-\alpha_{\min})
\exp\!\left[-\delta\odot\exp\!\left(\min(a,a_{\mathrm{cap}})\right)\right].
\end{gathered}
\label{eq:transition_coefficients}
\end{equation}
The hyperparameters satisfy
$0<\alpha_{\min}<\alpha_{\max}<1$, and the numerical cap
$a_{\mathrm{cap}}$ limits the exponent elementwise. The exponential factor is
related to the diagonal transition parameterization studied in
Parcae~\citep{prairie2026parcae}; the affine mapping sets the retention interval.
The learned coefficients are shared across sequences, positions, and iterations. With a
learned input projection $B$ and recurrent output normalization
$N_{\mathrm r}$, the state update is
\begin{equation}
 s^{(k)}=N_{\mathrm r}\!\left(
 R_\theta\!\left(\alpha\odot s^{(k-1)}+\delta\odot Be\right)
 \right),\qquad k=1,\ldots,K.
\label{eq:recursive}
\end{equation}
Here $\odot$ denotes channelwise multiplication broadcast over positions, and
$B$ acts on each position's feature vector. The term
$\alpha\odot s^{(k-1)}$ retains a channelwise fraction of the preceding state,
whereas $\delta\odot Be$ makes the input-conditioned prelude representation
directly available at every iteration. Each direct retention coefficient lies
strictly below one; the nonlinear core and normalization determine the
dynamics of the full update.

Normalization acts at two complementary boundaries. Sandwich RMS
normalization regulates the attention and feed-forward residual branches
within each core execution~\citep{geiping2025scaling}; $N_{\mathrm r}$
normalizes the state handed to the next iteration. The Transformer blocks use
bidirectional grouped-query attention~\citep{ainslie2023gqa}, rotary positional
embeddings~\citep{su2024roformer}, and SwiGLU layers~\citep{shazeer2020glu}.
Appendix~\ref{app:architecture} specifies the block equations and
initialization.

For masked reconstruction at depth $k$, a learned projection $W_{\mathrm o}$
maps the recurrent state into the coda, followed by output RMS normalization
$N_{\mathrm o}$. Writing this map as
$\Psi_\Theta(s)=N_{\mathrm o}(C(W_{\mathrm o}s))$, the tied embedding matrix
produces nucleotide logits $g^{(k)}$ and probabilities
\begin{equation}
\begin{gathered}
 h^{(k)}=\Psi_\Theta(s^{(k)}),\qquad g^{(k)}=h^{(k)}E^\top,\\
 p_{\Theta,k}(x_t\mid\tilde x)
 =\operatorname{softmax}\!\left(g_t^{(k)}\right)_{x_t}.
\end{gathered}
\label{eq:readout}
\end{equation}
The coda and tied reconstruction head are shared across depths; only the
recurrent state supplied to this pretraining interface changes with $k$.

\subsection{Training objective}
\label{sec:method_objective}

We train the shared update to produce useful predictions across inference
budgets. Each sequence $i$ independently samples a terminal depth $K_i$ and
receives masked-reconstruction supervision at that endpoint; the sampling law
is given in Section~\ref{sec:method_depth_training}. We couple this task
supervision to two criteria on the recurrent trajectory: relative displacement
regulates the final state update, and a one-sided reconstruction anchor
reinforces deep endpoints whose reconstruction loss exceeds a shallower
reference from the same trajectory. The combined objective is
\begin{equation}
\mathcal L=\mathcal L_{\mathrm{MLM}}
 +\lambda_{\mathrm{fp}}\mathcal L_{\mathrm{fp}}
 +\lambda_{\mathrm a}\mathcal L_{\mathrm{anchor}},
\label{eq:training_objective}
\end{equation}
where $\lambda_{\mathrm{fp}},\lambda_{\mathrm a}\geq0$ weight the auxiliary terms.

\paragraph{Terminal reconstruction.}
For a sequence $i$ with supervised positions $\mathcal M_i\ne\varnothing$, define its reconstruction loss at depth $k$ as
\begin{equation}
\ell_i^{(k)}=-\frac{1}{|\mathcal M_i|}
 \sum_{t\in\mathcal M_i}\log p_{\Theta,k}(x_{it}\mid\tilde x_i).
\label{eq:sequence_ce}
\end{equation}
Let $\mathcal I_{\mathrm M}$ be the supervised sequences in a microbatch and $Z_{\mathrm M}=\sum_{i\in\mathcal I_{\mathrm M}}|\mathcal M_i|$. The masked language modeling (MLM) objective~\citep{devlin2019bert} averages terminal reconstruction over supervised tokens:
\begin{equation}
\mathcal L_{\mathrm{MLM}}=
 \frac{1}{Z_{\mathrm M}}\sum_{i\in\mathcal I_{\mathrm M}}
 |\mathcal M_i|\ell_i^{(K_i)}.
\label{eq:mlm_loss}
\end{equation}
Sampling distributes terminal reconstruction supervision across depths over
sequences and training updates.

\paragraph{Relative terminal displacement.}
Motivated by terminal fixed-point regularization~\citep{viakhirev2026think},
we penalize the final recurrent update relative to the scale of the preceding
state. For attention-valid positions
$V_i$, let $\|u\|_{V_i}^2=\sum_{t\in V_i}\|u_t\|_2^2$. The relative
displacement penalty is
\begin{equation}
\mathcal L_{\mathrm{fp}}=
 \frac{1}{|\mathcal I_{\mathrm{fp}}|}
 \sum_{i\in\mathcal I_{\mathrm{fp}}}
 \frac{\|s_i^{(K_i)}-s_i^{(K_i-1)}\|_{V_i}^{2}}
 {\max\!\left(\|s_i^{(K_i-1)}\|_{V_i}^{2},
 \varepsilon_{\mathrm{fp}}\right)},
\label{eq:fixed_point_loss}
\end{equation}
where $\mathcal I_{\mathrm{fp}}=\{i:K_i\geq2\}$ and
$\varepsilon_{\mathrm{fp}}>0$ is a numerical floor. Per-sequence
normalization measures the terminal update relative to the scale of its
preceding state. The penalty covers all attention-valid positions, including
positions without reconstruction targets, and differentiates both terminal
states within the configured backward window.

The readout's local sensitivity relates terminal state displacement to the
corresponding change in logits (Appendix~\ref{app:objective_analysis}). Our
one-sided reconstruction anchor complements this state-level criterion by
acting directly on reconstruction loss.

\paragraph{One-sided reconstruction anchoring.}
We introduce a one-sided reconstruction anchor that uses a shallow reference
loss to determine when a deeper prediction receives additional supervision.
Let $K_{\mathrm a}$ be the
reference depth, $m_{\mathrm a}\geq0$ a tolerance, and
$\mathcal I_{\mathrm a}=\{i\in\mathcal I_{\mathrm M}:K_i>K_{\mathrm a}\}$
the supervised sequences whose terminal depth exceeds the reference. We define
\begin{equation}
\mathcal L_{\mathrm{anchor}}=
 \frac{1}{|\mathcal I_{\mathrm a}|}
 \sum_{i\in\mathcal I_{\mathrm a}}
 \left[\ell_i^{(K_i)}
 -\operatorname{stopgrad}\!\left(\ell_i^{(K_{\mathrm a})}\right)
 -m_{\mathrm a}\right]_+,
\label{eq:anchor_loss}
\end{equation}
where $[u]_+=\max(u,0)$ and $\operatorname{stopgrad}$ removes differentiation
through its argument. The reference and terminal losses use the same corrupted
input, initial state, supervised positions, and model parameters. The reference
state is captured from the same recurrent trajectory and read out without
gradient tracking.

The optimization effect of this one-sided form is explicit in its gradient.
Denote the per-sequence hinge in Equation~\eqref{eq:anchor_loss} by
$r_i^{\mathrm a}$. Along the retained training graph, its gradient is
\begin{equation}
\nabla_\Theta r_i^{\mathrm a}
=\mathbf 1\!\left\{\ell_i^{(K_i)}>\ell_i^{(K_{\mathrm a})}+m_{\mathrm a}\right\}
\nabla_\Theta\ell_i^{(K_i)}.
\label{eq:anchor_gradient}
\end{equation}
Here $\mathbf 1\{\cdot\}$ is the indicator function. When an eligible deep
endpoint's loss exceeds its detached reference by more than $m_{\mathrm a}$,
the anchor adds the gradient of that endpoint's reconstruction loss;
$\lambda_{\mathrm a}$ controls its contribution to
Equation~\eqref{eq:training_objective}. The observed nucleotides specify the
learning target, and the reference determines when to strengthen that
supervision. Improvements beyond the reference remain unpenalized, leaving
the representations and predictions free to evolve with depth.

MLM is averaged over supervised tokens, and each auxiliary objective over its
eligible sequences.
Appendix~\ref{app:training_algorithm} specifies empty-set conventions and
gradient paths.

\subsection{Depth sampling and truncated backpropagation}
\label{sec:method_depth_training}

We sample the terminal depth independently for every sequence, exposing the
shared update and readout to different computational budgets within a
microbatch~\citep{prairie2026parcae}. We draw a base depth from a
Poisson--lognormal distribution~\citep{geiping2025scaling} and impose a minimum
depth $K_{\min}$:
\begin{equation}
\begin{gathered}
\tau_i\sim\mathcal N\!\left(\log\mu_K-\frac{\sigma_K^2}{2},\sigma_K^2\right),\\
\widetilde K_i\sim\operatorname{Poisson}(e^{\tau_i}),
\qquad K_i=\max(K_{\min},\widetilde K_i).
\end{gathered}
\label{eq:depth_sampling}
\end{equation}
Here $\mu_K$ is the expected Poisson intensity before applying the lower bound
$K_{\min}$, and $\sigma_K$ controls dispersion in log intensity. This law
places most probability on shorter trajectories while retaining a tail of
larger compute budgets.

Forward depth and differentiated depth serve distinct roles. The sampled
$K_i$ determines the complete trajectory used for prediction, while a terminal
window $W_{\mathrm{bwd}}$ controls recurrent activation storage. All $K_i$
iterations execute in the forward pass, and gradients are retained through the
final $\min(K_i,W_{\mathrm{bwd}})$
iterations~\citep{geiping2025scaling,prairie2026parcae}. A prefix outside
this window is detached before the differentiated suffix begins.

The detached prefix still contributes its terminal state to the prediction;
gradient propagation is restricted to the retained suffix, where persistent
input injection also provides a path to the prelude. The configured suffix retains both
states required by the displacement objective, while the anchor readout remains
detached. Sampling $K_i$ before applying the backward window preserves the
distribution of forward budgets independently of this memory control.
Appendix~\ref{app:training_algorithm} gives the wavefront schedule and the
distributed normalization.

\subsection{Pretraining and inference}
\label{sec:method_training_protocol}

Stage~A pretrains the complete Eco-RNA backbone end to end from random
initialization with masked reconstruction.
Its data mixture is dominated by OpenGenome2 genomic sequences~\citep{brixi2026genome} and RNAcentral~\citep{rnacentral2019rnacentral}.
Stage~B continues from the Stage-A checkpoint on the same sources while shifting the sampling mixture toward RNA, without changing the architecture, objective, or prediction interface.
Downstream results use the Stage-B checkpoint, whereas the pretraining analysis in Section~\ref{sec:exp-pretraining} follows Stage-A checkpoints along training.
Appendix~\ref{app:data} describes the data sources and training configurations.

At inference, $K$ is specified explicitly. A fixed, position-indexed initial-state table is reused throughout each depth sweep, isolating the effect of additional core applications.
Fitness prediction uses the pretrained checkpoint directly.
Each supervised task starts from the same Stage-B pretrained backbone; after task adaptation, its checkpoint and task head are held fixed across $K$.
Section~\ref{sec:exp-overview} defines the task-specific readouts.

%% file: sections/experiments.tex
\subsection{Protocol}
\label{sec:exp-overview}

We begin with self-supervised reconstruction to establish whether repeated
computation is useful for the objective used to train Eco-RNA. We then test
whether this benefit extends to RNA fitness prediction at a competitive
computational cost. Assay-level analyses examine where the aggregate gains
arise. Finally, supervised tasks assess whether additional iterations remain
useful across prediction scales after task-specific fine-tuning.

Within each downstream depth sweep, we vary $K$ from 1 to 8 while holding
model weights, inputs, initial states, and any prediction heads fixed. Fitness
prediction uses the pretrained model without task-specific training and
follows the original 70-assay RNAGym protocol. Absolute Spearman correlation,
orientation-adjusted AUC, and absolute MCC are averaged within RNA type and
then equally across the five types. Family classification, chemical
reactivity, and secondary-structure prediction use separately fine-tuned
copies of the same backbone. Evaluation details appear in
Appendices~\ref{app:classification}--\ref{app:reactivity}.

\subsection{Progressive latent refinement emerges from pretraining}
\label{sec:exp-pretraining}

\begin{table}[!t]
\caption{\textbf{Reconstruction loss across pretraining and depth.}
Held-out RNA masked-reconstruction loss for Stage-A checkpoints at
$K{=}1,4,8$. Masked inputs and initial states are matched across depth within
each checkpoint; the final column reports the reduction from $K{=}1$ to 4.}
\label{tab:depth_pretraining}
\centering
\small
\setlength{\tabcolsep}{7pt}
\begin{tabular}{@{}ccccc@{}}
\toprule
\textbf{Pretraining updates} & $K{=}1$ & $K{=}4$ & $K{=}8$ & $\Delta_{1\to4}$ \\
\midrule
50k & 0.9357 & 0.8958 & 0.8955 & 0.040 \\
100k & 0.9070 & 0.8673 & 0.8667 & 0.040 \\
200k & 0.8829 & 0.8416 & 0.8408 & 0.041 \\
300k & 0.8772 & 0.8357 & 0.8348 & 0.042 \\
\bottomrule
\end{tabular}
\end{table}

We start with held-out reconstruction to assess whether repeated computation
helps Eco-RNA recover masked nucleotides during self-supervised pretraining.
We evaluate four Stage-A checkpoints between 50k and 300k updates, comparing
one, four, and eight iterations within each checkpoint
(Table~\ref{tab:depth_pretraining}). At each measured checkpoint, increasing
depth from one to four iterations reduces reconstruction loss by
0.040--0.042, while the reduction from four to eight iterations remains below
0.001. Later checkpoints have lower losses, but each also benefits from
additional iterations at inference. Thus, the shared update supports more
accurate reconstruction when applied repeatedly, and this benefit persists
throughout the measured pretraining trajectory. This establishes that extra
iterations are useful for the training objective before we assess their
value for downstream RNA prediction.

\subsection{Recurrent depth advances the compute-performance frontier}
\label{sec:exp-fitness}
\label{sec:exp-frontier}

We next evaluate whether the reconstruction gains are accompanied by better prediction of RNA function at a competitive computational cost.
RNA fitness prediction evaluates how well sequence-based scores rank experimentally measured variant effects.
RNAGym comprises 70 assays across five RNA types~\citep{arora2025rnagym}.
With model weights held fixed, increasing depth from one to four iterations raises aggregate absolute Spearman correlation from 0.242 to 0.286, an 18\%
relative gain, while MCC rises by 22\% (Appendix Table~\ref{tab:rnagym}).
A fixed-depth control matched in parameter count, block count, data, and training schedule reaches 0.224 absolute Spearman correlation.
Eco-RNA therefore exceeds this control after one iteration and improves further by repeating computation with the same learned weights.

To assess the cost of these gains, we estimate FLOPs for one forward pass on a 512-nucleotide input (Appendix~\ref{app:compute}).
Four Eco-RNA iterations require 2.0$\times$ the FLOPs of one iteration, while eight require 3.3$\times$.
Among the evaluated models, the $K{=}1$ through $4$ configurations lie on the Pareto frontier of estimated forward FLOPs and aggregate absolute Spearman correlation  (Figure~\ref{fig:pareto}).
At $K{=}4$, Eco-RNA exceeds Evo~2 40B on all three aggregate fitness metrics~\citep{brixi2026genome}, with 500$\times$ fewer parameters and approximately 200$\times$ fewer forward FLOPs.
One trained model therefore provides several competitive choices of prediction quality and computation, selected simply by changing the iteration count
(Appendix Table~\ref{tab:rnagym}).

\subsection{Assay-level analysis of depth gains}
\label{sec:exp-assays}

\begin{figure}[!htbp]
\centering
\includegraphics[width=\textwidth]{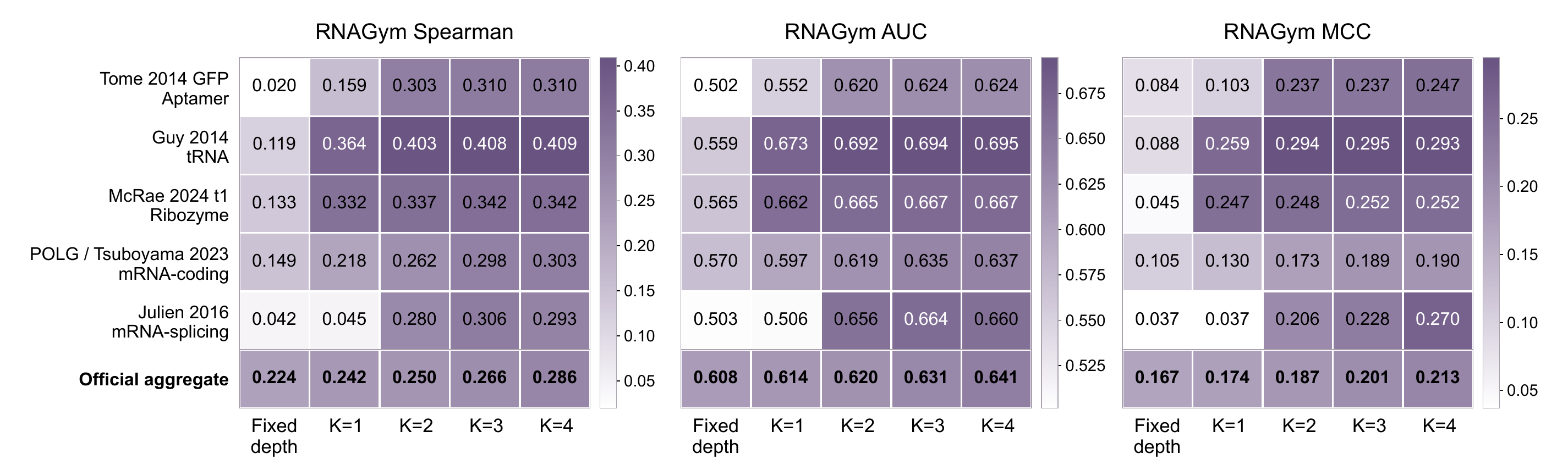}
\caption{\textbf{Depth-dependent fitness prediction on five selected assays.}
Columns: the fixed-depth control and $K{=}1,\ldots,4$
of one Eco-RNA checkpoint. Rows: one assay per RNA type; the last row (labeled
official aggregate) is the type-balanced aggregate over all 70 assays.}
\label{fig:rna_pertask}
\end{figure}

Additional depth produces different benefits across assays, even when the checkpoint and scoring rule remain fixed. Among the five selected examples in Figure~\ref{fig:rna_pertask}, the largest improvements occur in the splicing and aptamer assays, which also have the weakest first-iteration absolute Spearman correlations. A single additional iteration raises the splicing score from 0.045 to 0.280, showing that repeated application of the shared core can substantially improve variant ranking. The coding assay improves more gradually, whereas the tRNA and ribozyme examples begin with stronger predictions and show more modest gains.

The magnitude of a gain and the depth over which it develops are distinct aspects of refinement.
The aptamer assay improves sharply and then levels off, while the coding assay continues improving across the displayed depths.
The preferred depth can also depend on the evaluation criterion within the same assay: splicing Spearman correlation and AUC peak at $K{=}3$, whereas absolute MCC continues improving at $K{=}4$.
These examples reveal both rapid correction and gradual improvement under the same recurrent computation rule.
They motivate choosing depth according to the assay and the target metric, using validation to identify where additional iterations remain useful.
Adjustable depth makes this choice possible without retraining the model for each inference budget.

\subsection{Optimal execution depth scales with prediction granularity}

Finer-grained predictions retain benefits from refinement after sequence-level classification has largely saturated.
We examine this pattern using separately fine-tuned copies of Eco-RNA for family classification, per-nucleotide chemical reactivity, and pairwise secondary-structure prediction.
Within each task, the adapted backbone and prediction head remain fixed throughout the depth sweep.

Sequence-level classification reaches its best observed accuracy within a few iterations.
On the 13-class nRC benchmark, accuracy reaches 98.69\% at $K{=}4$ (Table~\ref{tab:depth_sweeps}).
At this reporting depth, Eco-RNA also leads the compared models  (Table~\ref{tab:transfer}).
The early plateau shows that additional iterations provide little further benefit for family assignment once this level of performance is reached.

\begin{table}[!htbp]
\caption{\textbf{RNA family classification and secondary-structure prediction.}
Models are fully fine-tuned from their pretrained checkpoints on each
benchmark. Eco-RNA reports its separately adapted task checkpoints at the
shared reporting depth of $K{=}4$;
column maxima are bolded.}
\label{tab:transfer}
\label{tab:beacon_ncrna}
\centering
\footnotesize
\setlength{\tabcolsep}{4pt}
\begin{tabular}{@{}lccc@{}}
\toprule
\textbf{Model} & \textbf{Params} & \textbf{ncRNA accuracy (\%)}$\uparrow$ & \textbf{SSP F1 (\%)}$\uparrow$ \\
\midrule
RNA-MSM & 96M & 84.85 & 57.98 \\
3UTRBERT & 87M & 94.32 & 59.41 \\
RNA-FM & 100M & 96.88 & 68.50 \\
RiNALMo & 150M & 91.89 & 75.02 \\
RNAGenesis & 866M & 97.81 & 73.72 \\
\midrule
\textbf{Eco-RNA ($K{=}4$)} & 80M & \textbf{98.69} & \textbf{75.40} \\
\bottomrule
\end{tabular}
\end{table}

Per-nucleotide reactivity shows a longer tail of improvement.
After adaptation on Ribonanza~\citep{he2024ribonanza}, we evaluate DMS and 2A3 predictions on 114{,}836 held-out RNAGym sequences~\citep{arora2025rnagym} (Appendix~\ref{app:reactivity}).
Eco-RNA leads all three probe-mean metrics at $K{=}4$ (Table~\ref{tab:rna_2d_structure}), while smaller gains persist to $K{=}7$, where all three means reach their best observed values (Table~\ref{tab:reactivity_full_depth}).
Thus, nucleotide-level prediction continues to benefit from refinement at depths beyond the classification plateau, with improvements in both discrimination and
continuous reactivity estimation.

\begin{table}[!htbp]
\caption{\textbf{Chemical-reactivity prediction.} Per-nucleotide AUC and
macro-F1 use global-median binarization; MAE is computed on continuous DMS and
2A3 reactivity, and Mean weights the two probes equally. We fine-tune the
learnable baselines~\citep{chen2022interpretable,fu2022ufold,sato2021mxfold2}
on Ribonanza under a shared supervised-adaptation protocol and
evaluate the structure-based methods~\citep{waymentsteele2022eternafold,
do2006contrafold,lorenz2011viennarna,reuter2010rnastructure} through the same
pipeline. Eco-RNA reports $K{=}4$; the complete sweep and
adaptation details appear in Table~\ref{tab:reactivity_full_depth} and
Appendix~\ref{app:reactivity}. Best displayed scores are in \textbf{bold}.}
\label{tab:rna_2d_structure}
\centering
\small
\setlength{\tabcolsep}{4pt}
\begin{tabular}{@{}l ccc ccc ccc@{}}
\toprule
 & \multicolumn{3}{c}{\textbf{DMS}} & \multicolumn{3}{c}{\textbf{2A3}} & \multicolumn{3}{c}{\textbf{Mean}} \\
\cmidrule(lr){2-4}\cmidrule(lr){5-7}\cmidrule(lr){8-10}
\textbf{Method} & \textbf{AUC}$\uparrow$ & \textbf{F1}$\uparrow$ & \textbf{MAE}$\downarrow$ & \textbf{AUC}$\uparrow$ & \textbf{F1}$\uparrow$ & \textbf{MAE}$\downarrow$ & \textbf{AUC}$\uparrow$ & \textbf{F1}$\uparrow$ & \textbf{MAE}$\downarrow$ \\
\midrule
RNAstructure & 0.646 & 0.605 & 0.326 & 0.721 & 0.667 & 0.300 & 0.683 & 0.636 & 0.313 \\
RNAfold (ViennaRNA) & 0.645 & 0.604 & 0.324 & 0.725 & 0.670 & 0.296 & 0.685 & 0.637 & 0.310 \\
CONTRAfold & 0.659 & 0.608 & 0.372 & 0.740 & 0.684 & 0.330 & 0.700 & 0.646 & 0.351 \\
EternaFold & 0.647 & 0.598 & 0.360 & 0.760 & 0.702 & 0.309 & 0.703 & 0.650 & 0.335 \\
\midrule
UFold & 0.841 & 0.766 & 0.182 & 0.727 & 0.668 & 0.244 & 0.784 & 0.717 & 0.213 \\
MXfold2 & 0.843 & 0.766 & 0.187 & 0.730 & 0.668 & 0.234 & 0.787 & 0.717 & 0.210 \\
RNA-FM & \textbf{0.853} & \textbf{0.780} & 0.173 & 0.799 & 0.722 & 0.201 & 0.826 & 0.751 & 0.187 \\
\textbf{Eco-RNA ($K{=}4$)} & \textbf{0.853} & 0.774 & \textbf{0.156} & \textbf{0.874} & \textbf{0.791} & \textbf{0.160} & \textbf{0.864} & \textbf{0.782} & \textbf{0.158} \\
\bottomrule
\end{tabular}
\end{table}

\begin{figure}[!htbp]
\centering
\includegraphics[width=\textwidth]{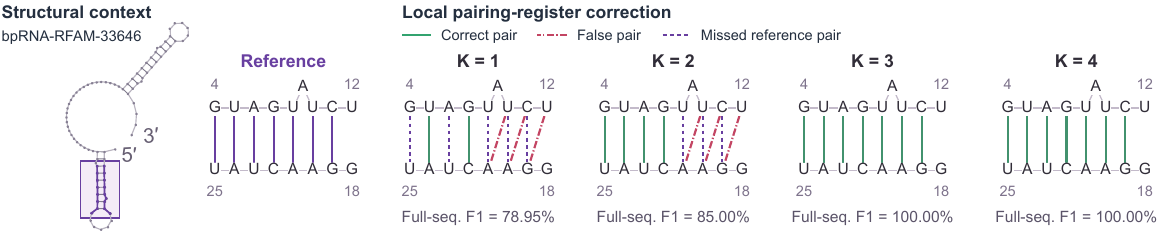}
\caption{\textbf{Local pairing refinement across recurrent depth.}
The left panel locates the displayed stem in bpRNA-RFAM-33646; the remaining
panels compare the reference with predictions at $K{=}1,\ldots,4$.
F1 scores are computed over the full sequence; the structural views show only the highlighted region.}
\label{fig:ssp_case_study}
\end{figure}

Pairwise structure prediction links the benefit of refinement to recovery of missing relationships. After fine-tuning on the bpRNA training split and evaluation on TS0~\citep{danaee2018bprna,singh2019spotrna}, most of the F1 gain appears early, followed by smaller improvements at greater depths (Table~\ref{tab:depth_sweeps}). From $K{=}1$ to $K{=}4$, recall increases by 3.3 percentage points while precision remains around 83--84\%, indicating that refinement recovers additional reference pairs at sustained precision. Eco-RNA exceeds the compared models at $K{=}4$ (Table~\ref{tab:transfer}). Figure~\ref{fig:ssp_case_study} illustrates a concrete correction in bpRNA-RFAM-33646: the local pairing-register error is resolved by $K{=}3$, and the corrected stem is retained at $K{=}4$. This example shows how additional iterations can revise specific pairing assignments. Appendix Figure~\ref{fig:ssp_case_study_full} provides further structural examples.

Together, these results suggest that the saturation of refinement gains is related to prediction granularity. A family label summarizes an entire sequence in one categorical decision, whereas nucleotide-level and pairwise outputs require accurate predictions at many positions or position pairs. Their continued improvement is consistent with additional iterations refining distinctions that matter for dense outputs after the benefits for coarse classification have plateaued. Across these evaluated tasks, coarse labels reach their performance plateau early, while finer-grained outputs retain smaller opportunities for correction at greater depth.

%% file: sections/discussion.tex
Eco-RNA establishes internal inference computation as a scaling dimension for RNA foundation models.
A single recurrent checkpoint provides several compute--performance operating points and advances the RNAGym Pareto frontier.
The ability to improve predictions by reusing learned parameters gives model users direct control over execution depth after training.
This complements the capacity and data choices made during pretraining and extends the computational flexibility of RNA sequence models.

The results connect this inference capability to both representation learning and downstream prediction.
Additional depth reduces held-out reconstruction loss at multiple stages of pretraining, and depth-dependent gains persist after adaptation to sequence labels, nucleotide pairs, and chemical reactivity.
The tasks differ in their response to additional computation: classification saturates early, whereas structure and reactivity show smaller gains at higher budgets.
For secondary structure, increased recall at sustained precision accounts for most of the improvement.
These observations establish the value of evaluating computational depth for each prediction task, with a common backbone supporting several useful inference budgets.

The training formulation links recurrent state dynamics to a criterion in prediction space.
Relative terminal displacement regularizes state movement, while the one-sided reconstruction anchor directs additional supervision toward deeper endpoints that underperform a reference depth.
Because the anchor retains the observed nucleotide as its target, improvements beyond the reference remain unpenalized.
The reference identifies depth-related degradation, while the reconstruction target specifies the desired prediction.
This separation provides a task-level criterion for training additional recurrent computation within a shared sequence model.

%% file: sections/statements.tex
\section*{Reproducibility statement}

Section~\ref{sec:method} defines the recurrent model and training objectives. Appendices~\ref{app:architecture}--\ref{app:optimization} specify the architecture, gradient paths, normalization, data sources, and pretraining configuration. Appendices~\ref{app:classification}, \ref{app:fitness}, and~\ref{app:reactivity} describe task scoring and adaptation, and Appendix~\ref{app:compute} derives the inference-compute estimate. Training and evaluation use published sequence and benchmark resources, including OpenGenome2, RNAcentral, RNAGym, and Ribonanza~\citep{brixi2026genome,rnacentral2019rnacentral,arora2025rnagym,he2024ribonanza}. Model weights, inference code, and evaluation configurations will be released publicly at a later date.

%% file: sections/appendix.tex
\numberwithin{equation}{section}
\renewcommand{\theHequation}{appendix.\thesection.\arabic{equation}}

\section{Method and evaluation details}
\label{app:details}

\subsection{Architecture and numerical specification}
\label{app:architecture}

During training, the Gaussian state in
Equation~\eqref{eq:state_initialization} is sampled independently and zeroed at
padding positions. Each evaluation sweep instead reuses one seed-fixed,
position-indexed Gaussian table at every depth, keeping initialization
invariant as $K$ changes.

The retention coefficient evaluates the exponent as
\(\exp(\min(a,a_{\mathrm{cap}}))\), bounding the numerical transition rate.
Transition vectors and boundary maps are shared across all iterations.

Each Transformer block combines bidirectional grouped-query attention with QK
normalization, RoPE, SwiGLU, and sandwich-normalized residual updates
\citep{ainslie2023gqa,su2024roformer,shazeer2020glu,geiping2025scaling}.
Writing \(N_j\) for RMS normalization, the block computes
\begin{equation}
\widehat u=N_2\!\left(u+\operatorname{Attn}(N_1(u))\right),\qquad
u'=N_4\!\left(\widehat u+\operatorname{MLP}(N_3(\widehat u))\right).
\label{app:sandwich_block}
\end{equation}
Embedding and linear weights use a zero-mean Gaussian initializer, with a
residual-scaled variant for attention and MLP output projections. Biases are
initialized to zero, RMSNorm gains to one, and \(B\) and \(W_{\mathrm o}\) to
the identity. Input embeddings and terminal vocabulary logits share the same
matrix. Table~\ref{tab:ed_model_specs} gives the complete backbone
configuration. Its \(6+3K\) executed blocks count one terminal path and exclude
the detached anchor readout used only during training.

\begin{table}[htbp]
\caption{\textbf{Eco-RNA backbone.}}
\label{tab:ed_model_specs}
\centering
\small
\begin{tabularx}{\textwidth}{@{}>{\raggedright\arraybackslash}p{0.39\textwidth}>{\raggedright\arraybackslash}X@{}}
\toprule
\textbf{Setting} & \textbf{Value} \\
\midrule
Trainable backbone parameters & 80,856,192 \\
Prelude / recurrent core / coda & \(3/3/3\) Transformer blocks \\
Distinct / executed blocks & \(9\) / \(6+3K\) \\
Hidden / feed-forward width & \(768/3{,}072\) \\
Query / key--value heads; head width & \(12/6\); \(64\) \\
Context length / active vocabulary & \(8{,}192\) tokens / \(23\) tokens \\
Tokenization & Single-nucleotide IUPAC; distinct DNA T and RNA U \\
Normalization / position encoding & RMSNorm (\(\epsilon=10^{-6}\)); RoPE base \(10^6\) \\
Initial state / evaluation state & \(\sigma_{\mathrm{init}}=1\); position table with seed 0 \\
Retention bounds / exponent cap & \(\alpha_{\min}=0.05,\ \alpha_{\max}=0.95\); \(a_{\mathrm{cap}}=30\) \\
Weight initialization & Gaussian standard deviation \(0.02\); residual output scale \(0.06455\) \\
Boundary maps / prediction head & \(B,W_{\mathrm o}\) identity-initialized; input/output embeddings tied \\
\bottomrule
\end{tabularx}
\end{table}

\subsection{Variable-depth training and objectives}
\label{app:training_algorithm}

\paragraph{Right-aligned execution and differentiation.}
Terminal depths \(K_i\) are sampled independently as in
Section~\ref{sec:method_depth_training}. Let \(K_{\max}=\max_iK_i\). At
wavefront step \(t\in\{1,\ldots,K_{\max}\}\), the active rows are
\begin{equation}
\mathcal A_t=\{i:t>K_{\max}-K_i\}.
\label{app:wavefront_set}
\end{equation}
Each row enters the wavefront late enough to complete exactly \(K_i\)
iterations at the common endpoint. All iterations execute forward. When
\(K_i>W_{\mathrm{bwd}}\), the first \(K_i-W_{\mathrm{bwd}}\) iterations omit
the autograd graph, and the boundary state is detached before differentiating
the final \(W_{\mathrm{bwd}}\) iterations. Shorter trajectories are
differentiated in full.

\begin{samepage}
Each training update proceeds as follows:
\begin{enumerate}
\item Corrupt the input and compute \(e_i\); sample \(s_i^{(0)}\) and \(K_i\).
\item Execute the right-aligned recurrent wavefront, retaining the terminal and
penultimate states and, when reached, the state at anchor depth \(K_{\mathrm a}\).
\item Read the terminal state with gradient tracking. Read a detached anchor
state through the same coda and tied head under no-gradient evaluation.
\item Form the masked-token and auxiliary objectives below, accumulate their
token-normalized gradients, and update the shared parameters.
\end{enumerate}
\end{samepage}

\paragraph{Supervision and corruption.}
Let \(V_i\) be the attention-valid positions, \(U_i\subseteq V_i\) the
non-special positions eligible for corruption, and
\(\mathcal M_i\subseteq U_i\) the positions selected for MLM supervision.
Special tokens participate in attention and state regularization but not in
MLM. The terminal reconstruction loss and its supervised-token normalization
are defined in Section~\ref{sec:method_objective}.

For span corruption, let $p_{\mathrm{sel}}$ be the target selection
fraction and $\mu_{\mathrm{span}}$ the exponential span scale. A nonempty row requests
\(q_i=\max\{1,\min(\lfloor p_{\mathrm{sel}}|U_i|\rfloor,|U_i|)\}\)
unique positions. Randomly placed, boundary-clipped spans with lengths
\(\max\{1,\lfloor Y\rfloor\}\), \(Y\sim
\operatorname{Exp}(1/\mu_{\mathrm{span}})\), are unioned until this count is
reached; \(q_i=0\) when \(U_i\) is empty. Conditional on selection, the
replacement rule in Tables~\ref{tab:ed_training} and~\ref{tab:ed_training_b}
is applied independently at each selected position. All selected positions
contribute to reconstruction.

The state-displacement term uses
\(\mathcal I_{\mathrm{fp}}=\{i:K_i\geq2\}\). For these sequences, define
\begin{equation}
r_i^{\mathrm{fp}}=
\frac{\|s_i^{(K_i)}-s_i^{(K_i-1)}\|_{V_i}^2}
{\max\{\|s_i^{(K_i-1)}\|_{V_i}^2,\varepsilon_{\mathrm{fp}}\}},
\qquad
\mathcal L_{\mathrm{fp}}=
\frac{1}{|\mathcal I_{\mathrm{fp}}|}
\sum_{i\in\mathcal I_{\mathrm{fp}}}r_i^{\mathrm{fp}},
\label{app:fixed_point_rows}
\end{equation}
where $\varepsilon_{\mathrm{fp}}>0$ is the numerical floor.
Anchor eligibility is
\(\mathcal I_{\mathrm a}=\{i:K_i>K_{\mathrm a},
|\mathcal M_i|>0\}\). The captured anchor state is detached, and its output
projection, coda, and tied head execute under no-gradient evaluation. An
auxiliary mean is defined as zero when its eligible set is empty; an optimizer
update requires a positive accumulated supervised-token count. At hinge
equality, the implementation selects the zero subgradient.

\paragraph{Relative displacement and readout change.}
\label{app:objective_analysis}
Fix the attention mask and view $\Psi_\Theta$ as a map over valid-position
representations. If its derivative has operator norm at most $L_{\Psi,i}$
along the segment from $s_i^{(K_i-1)}$ to $s_i^{(K_i)}$, integrating the
derivative along this segment and applying the tied embedding matrix gives
\begin{equation}
\bigl\|g_i^{(K_i)}-g_i^{(K_i-1)}\bigr\|_{V_i}
\leq \|E\|_2 L_{\Psi,i}
\bigl\|s_i^{(K_i)}-s_i^{(K_i-1)}\bigr\|_{V_i}.
\label{eq:readout_sensitivity}
\end{equation}
Here $\|E\|_2$ is the spectral norm of the embedding matrix. Define the
preceding-state scale
\begin{equation}
\begin{gathered}
\zeta_i=
\sqrt{\max\!\left\{\|s_i^{(K_i-1)}\|_{V_i}^2,
\varepsilon_{\mathrm{fp}}\right\}},\\
\|s_i^{(K_i)}-s_i^{(K_i-1)}\|_{V_i}
=\zeta_i\sqrt{r_i^{\mathrm{fp}}}.
\end{gathered}
\label{app:relative_displacement_identity}
\end{equation}
Substituting Equation~\eqref{app:relative_displacement_identity} yields the
corresponding relative relation
\begin{equation}
\frac{\|g_i^{(K_i)}-g_i^{(K_i-1)}\|_{V_i}}
{\zeta_i}
\leq \|E\|_2L_{\Psi,i}\sqrt{r_i^{\mathrm{fp}}}.
\label{app:readout_change_bound}
\end{equation}
These bounds relate relative terminal displacement to local changes in the
readout. The reconstruction anchor complements this state-level criterion by
acting directly on task loss through Equation~\eqref{eq:anchor_gradient}.

\paragraph{Accumulation and DDP normalization.}
\label{app:normalization}
Dynamic token batches are normalized over all supervised tokens across ranks
and accumulation microbatches. Let \(S_{rb}\) and \(m_{rb}\) denote summed
token cross-entropy and supervised-token count on rank \(r\), microbatch \(b\);
let \(Q\) be the world size and
\(Z_{\mathrm{acc}}=\sum_{r,b}m_{rb}\). With
\(A_{rb}=\lambda_{\mathrm{fp}}\mathcal L_{\mathrm{fp},rb}
+\lambda_{\mathrm a}\mathcal L_{\mathrm{anchor},rb}\), the update uses
\begin{equation}
\begin{gathered}
\mathcal L_{rb}^{\mathrm{model}}
=\frac{S_{rb}}{Z_{\mathrm{acc}}}
+\frac{m_{rb}}{Z_{\mathrm{acc}}}A_{rb},\\
\mathcal L^{\mathrm{update}}
=\frac{1}{Q}\sum_{r=1}^{Q}Q\sum_b\mathcal L_{rb}^{\mathrm{model}}
=\frac{1}{Z_{\mathrm{acc}}}\sum_{r,b}(S_{rb}+m_{rb}A_{rb}).
\end{gathered}
\label{app:distributed_normalization}
\end{equation}
The factor \(Q\) cancels DDP's rank average, while weighting each local
auxiliary mean by \(m_{rb}\) aligns it with that microbatch's contribution to
supervised-token count.

\subsection{Pretraining data and configuration}
\label{app:data}
\label{app:optimization}

Stage~A trains the complete Eco-RNA backbone from random initialization on
unlabeled sequences from four length-partitioned OpenGenome2 genomic sources,
its non-coding-RNA partition, and a filtered RNAcentral
corpus~\citep{brixi2026genome,rnacentral2019rnacentral}. Its schedule
progressively increases RNA sampling while retaining genomic exposure. DNA and
RNA sources use T and U, respectively, and sequences exceeding the context
limit are randomly cropped. Table~\ref{tab:ed_training} reports the Stage-A
configuration.

Stage~B continues self-supervised pretraining with greater exposure to RNA,
using the architecture and objective established in Stage~A.
Table~\ref{tab:ed_training_b} reports its configuration.

\begin{table}[htbp]
\caption{\textbf{Stage-A pretraining configuration.}}
\label{tab:ed_training}
\centering
\small
\begin{tabularx}{\textwidth}{@{}>{\raggedright\arraybackslash}p{0.39\textwidth}>{\raggedright\arraybackslash}X@{}}
\toprule
\textbf{Setting} & \textbf{Value} \\
\midrule
Data & OpenGenome2 genomic buckets and non-coding-RNA partition, filtered RNAcentral \\
Updates / maximum length & \(300{,}000\) / \(8{,}192\) input tokens \\
Position selection & \(p_{\mathrm{sel}}=0.15\); exponential span scale \(\mu_{\mathrm{span}}=5\) \\
Replacement after selection & 80\% mask / 10\% modality-valid random base / 10\% unchanged \\
Depth distribution & \(\mu_K=4,\ \sigma_K=0.5,\ K_{\min}=2\) \\
Backward window & \(W_{\mathrm{bwd}}=2\) terminal iterations \\
Displacement / anchor & \(\lambda_{\mathrm{fp}}=0.05,\ \lambda_{\mathrm a}=0.10\) \\
Anchor depth / margin & \(K_{\mathrm a}=4,\ m_{\mathrm a}=0\) \\
Fixed-point floor & \(\varepsilon_{\mathrm{fp}}=10^{-12}\) \\
Optimizer & AdamW, \((\beta_1,\beta_2)=(0.9,0.999)\), \(\epsilon=10^{-8}\) \\
Weight decay / clipping & \(0.1\) / global norm \(1.0\) \\
Learning-rate schedule & 15,000-update warmup to \(8\times10^{-5}\), then cosine; floor \(8\times10^{-6}\) \\
Training system & 8 NVIDIA H200 GPUs, BF16 DDP, activation checkpointing \\
\bottomrule
\end{tabularx}
\end{table}

\begin{table}[htbp]
\caption{\textbf{Stage-B pretraining configuration.}}
\label{tab:ed_training_b}
\centering
\small
\begin{tabularx}{\textwidth}{@{}>{\raggedright\arraybackslash}p{0.39\textwidth}>{\raggedright\arraybackslash}X@{}}
\toprule
\textbf{Setting} & \textbf{Value} \\
\midrule
Initialization & Stage-A checkpoint (300,000 updates) \\
Data & OpenGenome2 non-coding-RNA partition and RNAcentral at increased sampling weight; genomic buckets retained \\
Updates / maximum length & \(200{,}000\) / \(8{,}192\) input tokens \\
Position selection & Span masking; \(p_{\mathrm{sel}}=0.15\), exponential span scale \(\mu_{\mathrm{span}}=4\) \\
Replacement after selection & 80\% mask / 10\% random nucleotide / 10\% unchanged \\
Depth distribution / backward window & \(\mu_K=4,\ \sigma_K=0.5,\ K_{\min}=2\); \(W_{\mathrm{bwd}}=2\) \\
Displacement / anchor & \(\lambda_{\mathrm{fp}}=0.05,\ \lambda_{\mathrm a}=0.10\) \\
Anchor depth / margin & \(K_{\mathrm a}=4,\ m_{\mathrm a}=0\) \\
Optimizer & Fused AdamW, \((\beta_1,\beta_2)=(0.9,0.999)\), \(\epsilon=10^{-8}\) \\
Weight decay / clipping & \(0.1\) / global norm \(1.0\) \\
Learning-rate schedule & 5,400-update warmup to \(8\times10^{-5}\), then cosine; floor \(8\times10^{-6}\) \\
Training system & 8 NVIDIA H200 GPUs; BF16 DDP, activation checkpointing \\
\bottomrule
\end{tabularx}
\end{table}

\subsection{Compute accounting}
\label{app:compute}

Let \(L\) be sequence length, \(d\) the hidden width, \(d_{\mathrm{kv}}\) the
total key/value width, \(d_{\mathrm f}\) the feed-forward width, and
\(|\mathcal V|\) the vocabulary size. Counting a fused multiply--add as two
floating-point operations, the leading matrix-multiplication cost of one dense
bidirectional Transformer block is
\begin{equation}
F_{\mathrm{blk}}(L)=
4Ld(d+d_{\mathrm{kv}})+4L^2d+6Ldd_{\mathrm f}.
\label{app:compute_block}
\end{equation}
The input projection \(B e\) is recomputed at each iteration, \(W_{\mathrm o}\) is
applied once, and the tied vocabulary head is applied once. The resulting
single-forward count is
\begin{equation}
F_{\mathrm{Eco}}(K,L)=
D(K)F_{\mathrm{blk}}(L)+2(K+1)Ld^2+2Ld|\mathcal V|.
\label{app:compute_total}
\end{equation}
For \(L=512\), the count is 90.011, 180.608, and 301.404 GFLOPs at
\(K=1,4,8\); it covers the dense matrix multiplications of one forward pass,
which dominate cost at this scale. The compared models in
Figure~\ref{fig:pareto} are counted the same way from their released
checkpoints and inference code.

\subsection{Non-coding RNA classification and secondary-structure prediction}
\label{app:classification}
\label{app:ssp}

Family classification follows the published train, validation, and test split
of the 13-class nRC benchmark~\citep{fiannaca2017nrc,rossi2019ncrna}. SSP uses
the bpRNA training split and TS0 test set
\citep{danaee2018bprna,singh2019spotrna}. Both tasks adapt the full Eco-RNA
backbone with a task-specific head: a lightweight classifier over the terminal
\texttt{[CLS]} representation for nRC and a pairwise head over terminal coda
features for SSP. SSP reports strict global full-matrix micro-\(F_1\), pooling
valid cells of the complete \(L\times L\) matrices after directly thresholding
pair probabilities at 0.5. Within each task, the adapted
checkpoint and head remain fixed throughout the \(K=1,\ldots,8\) sweep.

\begin{figure}[p]
\centering
\includegraphics[width=\textwidth]{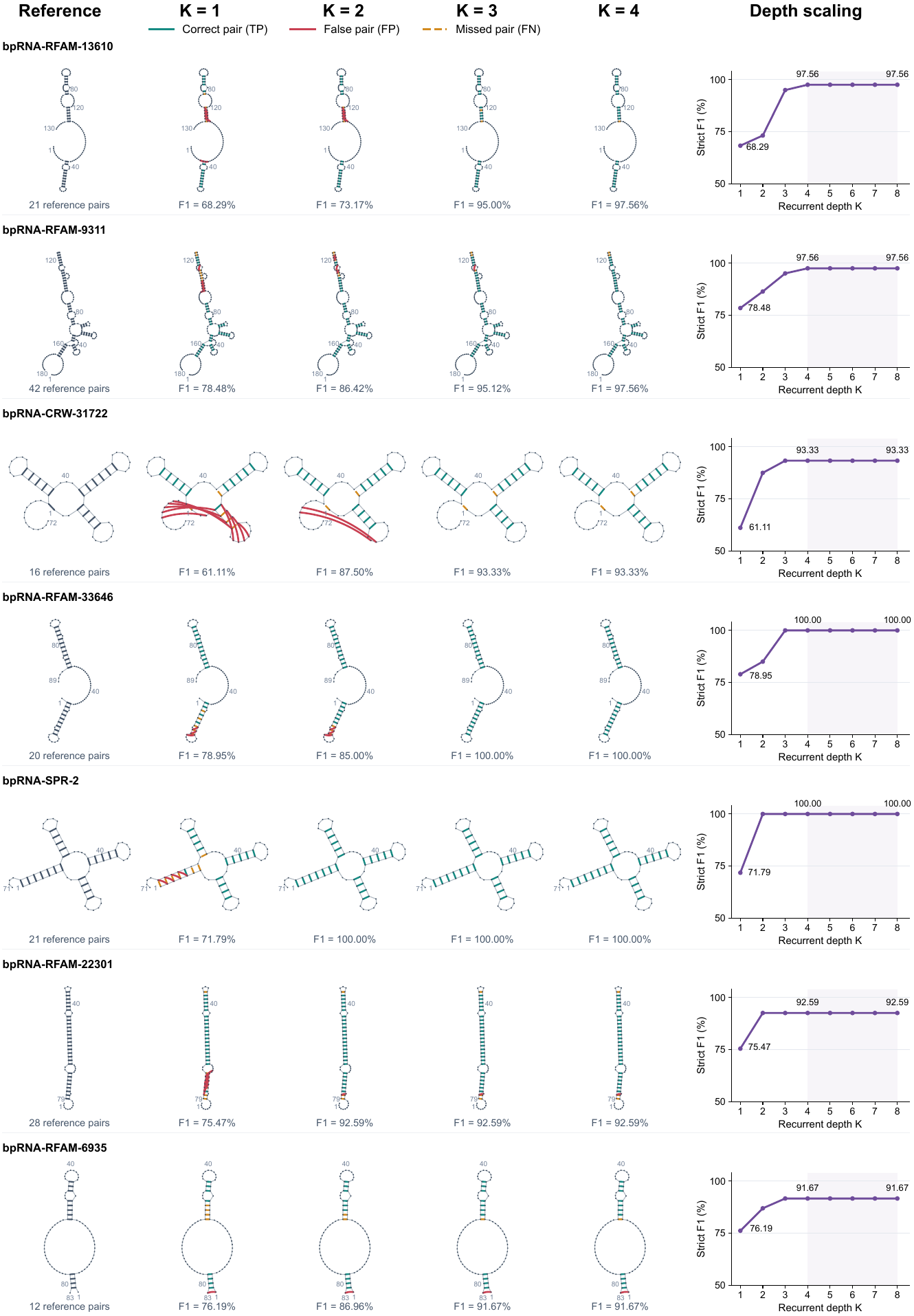}
\caption{\textbf{Secondary-structure case studies across recurrent depths.}
Rows show seven selected bpRNA sequences, with reference structures,
Eco-RNA predictions at $K{=}1,\ldots,4$, and full-sequence strict F1 over
$K{=}1,\ldots,8$. Teal, red, and dashed orange edges denote correct, false,
and missed base pairs, respectively.}
\label{fig:ssp_case_study_full}
\end{figure}

\subsection{RNAGym fitness prediction}
\label{app:fitness}

Fitness is evaluated zero-shot on the 70 RNAGym deep mutational scanning
assays~\citep{arora2025rnagym}. Let \(w\) be an assay's wild-type sequence,
\(v\) a variant, and \(\mathcal D(v)\) its substituted positions. The
wild-type-marginal score uses the uncorrupted wild-type context and sums the
log-odds of the mutant nucleotide against the wild-type nucleotide at each
substitution:
\begin{equation}
F_K(v;w)=\sum_{j\in\mathcal D(v)}
\left[\log p_{\Theta,K}(v_j\mid w,j)
-\log p_{\Theta,K}(w_j\mid w,j)\right].
\label{app:fitness_score}
\end{equation}
Here \(p_{\Theta,K}(\cdot\mid w,j)\) is the vocabulary distribution at position
\(j\) under the wild-type input \(w\). Every substitution in a variant is
scored under this shared context.

For each assay \(a\), the measured-fitness vector $y$ is binarized above its median for AUC
and MCC, and predicted scores are binarized above their median for MCC. The
reported orientation-invariant assay metrics are
\begin{equation}
q_{a,\rho}=|\operatorname{Spearman}(F_K,y)|,\quad
q_{a,\mathrm{AUC}}=\max\{\mathrm{AUC}_a,1-\mathrm{AUC}_a\},\quad
q_{a,\mathrm{MCC}}=|\mathrm{MCC}_a|.
\label{app:fitness_metrics}
\end{equation}
Let \(\mathcal T\) contain the five RNA types: mRNA-coding, ribozyme, tRNA,
mRNA-splicing, and aptamer, and let \(\mathcal A_t\) be the assays of type
\(t\). Each headline metric \(m\in\{\rho,\mathrm{AUC},\mathrm{MCC}\}\) first
averages assays within type and then weights the five types equally:
\begin{equation}
Q_m=\frac{1}{|\mathcal T|}\sum_{t\in\mathcal T}
\frac{1}{|\mathcal A_t|}\sum_{a\in\mathcal A_t}q_{a,m}.
\label{app:fitness_aggregate}
\end{equation}
Checkpoint, alphabet convention, initial-state table, and scoring rule remain
fixed throughout each depth sweep.

The fixed-depth control matches the recurrent model's parameter count, block
count, data, and training schedule. It executes nine distinct blocks once per
sequence as a conventional masked language model and uses the same fitness
scoring and aggregation. Table~\ref{tab:rnagym} reports the model comparison
with the complete fitness depth sweep. Table~\ref{tab:depth_sweeps} reports
the corresponding sweeps after supervised adaptation.

\begin{table}[htbp]
\caption{\textbf{Zero-shot RNA fitness prediction across models and recurrent depths.}
Evaluation uses RNAGym~\citep{arora2025rnagym} with the type-balanced
aggregation described in Appendix~\ref{app:fitness}. Eco-RNA evaluates one
checkpoint at $K{=}1,\ldots,8$, with inputs, initial states, and scoring fixed;
the matched fixed-depth control omits recurrence. Baseline and control scores
retain three decimal places; recurrent Eco-RNA scores use four.
Bold marks column maxima before rounding.}
\label{tab:rnagym}
\label{tab:fitness_full_depth}
\centering
\footnotesize
\setlength{\tabcolsep}{4pt}
\begin{tabular}{@{}lcccc@{}}
\toprule
\textbf{Model} & \textbf{Params} & \textbf{Spearman}$\uparrow$ & \textbf{AUC}$\uparrow$ & \textbf{MCC}$\uparrow$ \\
\midrule
GenSLM~\citep{zvyagin2023genslms} & 2.5B & 0.122 & 0.558 & 0.083 \\
Evo~1~\citep{nguyen2024sequence} & 6.5B & 0.124 & 0.561 & 0.091 \\
RiNALMo~\citep{penic2025rinalmo} & 650M & 0.155 & 0.582 & 0.122 \\
Nucleotide Transformer~\citep{dalla2025nucleotide} & 2.5B & 0.163 & 0.577 & 0.115 \\
Evo~1.5~\citep{arora2025rnagym} & 6.5B & 0.177 & 0.583 & 0.131 \\
Nucleotide Transformer v3~\citep{boshar2025foundational} & 650M & 0.189 & 0.593 & 0.133 \\
RNAErnie~\citep{wang2024rnaernie} & 87M & 0.189 & 0.598 & 0.154 \\
AIDO.RNA~\citep{zou2024aido} & 1.6B & 0.203 & 0.596 & 0.135 \\
Evo~2 7B~\citep{brixi2026genome} & 7B & 0.214 & 0.604 & 0.161 \\
RNA-FM~\citep{chen2022interpretable} & 100M & 0.217 & 0.606 & 0.155 \\
Evo~2 40B~\citep{brixi2026genome} & 40B & 0.276 & 0.636 & 0.212 \\
\midrule
Eco-RNA, fixed-depth control & 80M & 0.224 & 0.608 & 0.167 \\
Eco-RNA ($K=1$) & 80M & 0.2424 & 0.6138 & 0.1741 \\
Eco-RNA ($K=2$) & 80M & 0.2501 & 0.6199 & 0.1870 \\
Eco-RNA ($K=3$) & 80M & 0.2659 & 0.6309 & 0.2007 \\
Eco-RNA ($K=4$) & 80M & \textbf{0.2861} & \textbf{0.6409} & \textbf{0.2130} \\
Eco-RNA ($K=5$) & 80M & 0.2857 & 0.6407 & 0.2108 \\
Eco-RNA ($K=6$) & 80M & 0.2858 & \textbf{0.6409} & 0.2083 \\
Eco-RNA ($K=7$) & 80M & 0.2851 & 0.6406 & 0.2059 \\
Eco-RNA ($K=8$) & 80M & 0.2854 & 0.6407 & 0.2074 \\
\bottomrule
\end{tabular}
\end{table}

\begin{table}[htbp]
\caption{\textbf{Fixed-weight depth sweeps after supervised adaptation.}
The checkpoint, task head, inputs, and initial states remain fixed within each
sweep. Classification and SSP are percentages; reactivity metrics average
the DMS and 2A3 probes.
Bold marks each row's best value before rounding.}
\label{tab:depth_sweeps}
\label{tab:transfer_depth}
\label{tab:classification_depth}
\label{tab:reactivity_depth}
\centering
\small
\setlength{\tabcolsep}{3.5pt}
\begin{tabular}{@{}l*{8}{c}@{}}
\toprule
$\boldsymbol{K}$ & 1 & 2 & 3 & 4 & 5 & 6 & 7 & 8 \\
\midrule
\multicolumn{9}{@{}l}{\textit{nRC family classification}} \\
Accuracy (\%)$\uparrow$ & 97.96 & 98.46 & 98.62 & \textbf{98.69} & 98.65 & 98.65 & \textbf{98.69} & \textbf{98.69} \\
\addlinespace
\multicolumn{9}{@{}l}{\textit{bpRNA TS0 secondary structure}} \\
$F_1$ (\%)$\uparrow$ & 73.10 & 75.10 & 75.38 & 75.40 & 75.47 & \textbf{75.48} & 75.47 & 75.48 \\
Precision (\%)$\uparrow$ & 83.00 & 83.66 & 83.70 & 83.67 & 83.69 & 83.69 & 83.69 & \textbf{83.70} \\
Recall (\%)$\uparrow$ & 65.31 & 68.13 & 68.57 & 68.62 & 68.72 & \textbf{68.74} & 68.72 & 68.72 \\
\addlinespace
\multicolumn{9}{@{}l}{\textit{RNAGym chemical reactivity, mean over probes}} \\
AUC$\uparrow$ & 0.8212 & 0.8500 & 0.8604 & 0.8636 & 0.8651 & 0.8651 & \textbf{0.8655} & 0.8654 \\
$F_1\uparrow$ & 0.7491 & 0.7715 & 0.7798 & 0.7825 & 0.7840 & 0.7841 & \textbf{0.7843} & 0.7842 \\
MAE$\downarrow$ & 0.1945 & 0.1685 & 0.1608 & 0.1580 & 0.1569 & 0.1567 & \textbf{0.1565} & 0.1565 \\
\bottomrule
\end{tabular}
\end{table}

\subsection{Chemical-reactivity prediction}
\label{app:reactivity}

The RNAGym secondary-structure track evaluates per-nucleotide DMS and 2A3
chemical reactivity~\citep{arora2025rnagym}. For our supervised evaluation,
we fine-tune learnable models on the Ribonanza chemical-mapping
corpus~\citep{he2024ribonanza} and evaluate on 114,836 held-out sequences.
We evaluate RNA-FM, UFold, and MXfold2~\citep{chen2022interpretable,
fu2022ufold,sato2021mxfold2} with matched reactivity heads; thermodynamic
predictors supply unpaired-probability reactivity scores from their structure
outputs.

Eco-RNA is adapted jointly with a four-block convolutional head whose input
concatenates the terminal representation and its difference from the
first-iteration representation; this difference is zero at \(K=1\). The
adapted checkpoint, head, and initial-state table remain fixed throughout the
\(K=1,\ldots,8\) sweep.

AUC and macro-\(F_1\) use global-median binarization, MAE uses continuous
reactivity, and each mean weights DMS and 2A3 equally.
Table~\ref{tab:rna_2d_structure} reports Eco-RNA at \(K=4\) alongside the baselines;
Tables~\ref{tab:depth_sweeps} and~\ref{tab:reactivity_full_depth} provide the
complete depth results as probe means and per-probe scores, respectively.

\begin{table}[htbp]
\caption{\textbf{Chemical-reactivity prediction across recurrent depths.}
The fine-tuned checkpoint and head are fixed across $K$.
Metrics follow Table~\ref{tab:rna_2d_structure}; Mean weights DMS and 2A3 equally.
Each column's best value is in \textbf{bold}.}
\label{tab:reactivity_full_depth}
\centering
\small
\setlength{\tabcolsep}{3pt}
\begin{tabular}{@{}c ccc ccc ccc@{}}
\toprule
 & \multicolumn{3}{c}{\textbf{DMS}} & \multicolumn{3}{c}{\textbf{2A3}} & \multicolumn{3}{c}{\textbf{Mean}} \\
\cmidrule(lr){2-4}\cmidrule(lr){5-7}\cmidrule(lr){8-10}
$\boldsymbol{K}$ & \textbf{AUC}$\uparrow$ & \textbf{F1}$\uparrow$ & \textbf{MAE}$\downarrow$ & \textbf{AUC}$\uparrow$ & \textbf{F1}$\uparrow$ & \textbf{MAE}$\downarrow$ & \textbf{AUC}$\uparrow$ & \textbf{F1}$\uparrow$ & \textbf{MAE}$\downarrow$ \\
\midrule
1 & 0.836217 & 0.767908 & 0.179242 & 0.806089 & 0.730326 & 0.209712 & 0.821153 & 0.749117 & 0.194477 \\
2 & 0.846327 & 0.771044 & 0.162480 & 0.853681 & 0.771873 & 0.174473 & 0.850004 & 0.771458 & 0.168476 \\
3 & 0.854592 & \textbf{0.776349} & 0.156537 & 0.866237 & 0.783168 & 0.165125 & 0.860415 & 0.779758 & 0.160831 \\
4 & 0.853233 & 0.773936 & 0.156149 & 0.873936 & 0.791049 & 0.159765 & 0.863584 & 0.782492 & 0.157957 \\
5 & \textbf{0.855257} & 0.775581 & \textbf{0.154953} & 0.875030 & 0.792345 & 0.158763 & 0.865144 & 0.783963 & 0.156858 \\
6 & 0.854139 & 0.774719 & 0.155325 & 0.876121 & 0.793491 & 0.158016 & 0.865130 & 0.784105 & 0.156670 \\
7 & 0.854830 & 0.775125 & 0.155065 & 0.876097 & 0.793524 & 0.157951 & \textbf{0.865464} & \textbf{0.784325} & \textbf{0.156508} \\
8 & 0.854335 & 0.774483 & 0.155280 & \textbf{0.876384} & \textbf{0.793841} & \textbf{0.157747} & 0.865359 & 0.784162 & 0.156514 \\
\bottomrule
\end{tabular}
\end{table}

%% file: sections/related_work.tex
\paragraph{RNA and nucleotide foundation models.}
Scaling nucleotide foundation models has primarily meant increasing pretraining
data, parameter capacity, or accessible context. RNA-FM~\citep{chen2022interpretable},
RiNALMo~\citep{penic2025rinalmo}, and AIDO.RNA~\citep{zou2024aido} learn
general-purpose representations from unannotated RNA, while Evo~2 extends
nucleotide modeling across genomic domains at substantially larger
scale~\citep{brixi2026genome}. DNA-focused models have likewise expanded the
design space through new tokenizations and sequence mixers, including
DNABERT-2~\citep{zhou2023dnabert}, Nucleotide
Transformer~\citep{dalla2025nucleotide}, HyenaDNA~\citep{nguyen2023hyenadna},
and Caduceus~\citep{schiff2024caduceus}. Across these architectures, the
pretrained backbone ordinarily executes a prescribed number of transformations
for every prediction. Eco-RNA introduces inference depth as a separate scaling
axis: one checkpoint supports multiple computational budgets while preserving
its input and prediction interface.

\paragraph{Recurrent depth and computation allocation.}
Test-time computation can expand an output trace or search over additional
candidates~\citep{brown2024large,snell2024scaling}; recurrent depth allocates
that computation to further transformation of the latent representation.
Universal Transformers established the repeated application of a shared
transition, optionally coupled to adaptive halting~\citep{Dehghani2018}, whereas
ALBERT shares parameters across a fixed number of layers to improve parameter
efficiency~\citep{lan2019albert}. Huginn studies recurrent-depth language
modeling with a shared middle core between a prelude and coda and an iteration
count that remains selectable after training~\citep{geiping2025scaling}. Other
designs allocate recurrence at different granularities. RINS repeats an early
network segment and combines
stochastic recursion with depth-indexed adapters~\citep{alabdulmohsin2025recursive};
Mixture-of-Recursions assigns computation at the token level~\citep{bae2025mixture};
and SpiralFormer couples a shared core to coarse-to-fine resolution
changes~\citep{yu2026spiralformer}. Eco-RNA applies a full-resolution sequence
update with a sequence-wide iteration count chosen at inference.

\paragraph{Training shared updates across depths.}
A reusable update must learn over multiple prediction horizons while remaining
tractable to optimize. Huginn combines Poisson--lognormal depth sampling with
backpropagation through a terminal suffix~\citep{geiping2025scaling}. Parcae
uses an input-driven diagonal state transition, samples terminal depth per
sequence, and separates executed forward depth from the retained backward
window~\citep{prairie2026parcae}. Complementary work examines the
dynamics of repeated parameter use: DeepLoop derives residual and initialization
scaling rules~\citep{li2026deeploop}, while STARS regularizes an estimate of the
recurrent Jacobian's spectral radius~\citep{yang2026stabilizing}. Think Shallow,
Solve Deep analyzes finite-time behavior through terminal state displacement and
latent anchoring~\citep{viakhirev2026think}.

Eco-RNA formulates cross-depth learning around the predictive role of the
recurrent state. Persistent input injection and bounded retention condition the
shared Transformer update, while normalization controls its scale. Relative
terminal displacement regulates state motion, and a one-sided reconstruction
anchor connects the trajectory to masked-token prediction. State matching and
output distillation supervise a target representation or distribution; the
reconstruction anchor uses shallow loss to gate additional supervision at the
deeper endpoint. The observed nucleotides remain the targets, leaving deeper
predictions free to improve beyond the reference.

\paragraph{Iterative computation in biological models.}
Biological prediction systems have also benefited from iteration, although the
recurrent object and learning objective differ across settings. RNAformer
recycles a two-dimensional pairing representation under supervised
secondary-structure training~\citep{franke2023scalable}, and RhoFold+ feeds a
predicted three-dimensional structure back through successive refinement
rounds~\citep{shen2024accurate}. In protein modeling, ESMFold2 applies looped
folding layers to pretrained sequence representations and evaluates additional
loops in selected settings~\citep{candido2026language}. GENA-LM instead carries
memory tokens between consecutive genomic segments during
fine-tuning~\citep{fishman2025gena}. Eco-RNA places recurrence within the
self-supervised sequence backbone, making adjustable inference depth available
across multiple RNA prediction tasks.